\documentclass[letterpaper, 10 pt, conference]{ieeeconf}  
\usepackage{url}
\usepackage{graphicx}
\usepackage{booktabs}
\usepackage{array}
\usepackage{tabularx}
\usepackage{caption}
\usepackage{subfig}

\IEEEoverridecommandlockouts                              

\title{\LARGE \bf
LeAD: The LLM Enhanced Planning System Converged with End-to-end Autonomous Driving
}

\author{Yuhang~Zhang,
        Jiaqi~Liu, 
        Chengkai~Xu,
        Peng~Hang,~\IEEEmembership{Senior Member,~IEEE,}
        and~Jian~Sun
\thanks{This work was supported in part by the National Natural Science Foundation of China (52302502), the State Key Laboratory of Intelligent Green Vehicle and Mobility under Project No. KFZ2408, the State Key Lab of Intelligent Transportation System under Project No. 2024-A002, and the Fundamental Research Funds for the Central Universities.}
\thanks{Y. Zhang, J. Liu, C. Xu, P. Hang and J. Sun are with the College of Transportation, Tongji University, Shanghai 201804, China. (e-mail: {2153338, liujiaqi13, 2151162, hangpeng, sunjian}@tongji.edu.cn)}
\thanks{Corresponding author: Peng Hang}
}

\begin{document}

\maketitle
\thispagestyle{empty}
\pagestyle{empty}

\begin{abstract}
A principal barrier to large-scale deployment of urban autonomous driving systems lies in the prevalence of complex scenarios and edge cases. Existing systems fail to effectively interpret semantic information within traffic contexts and discern intentions of other participants, consequently generating decisions misaligned with skilled drivers' reasoning patterns. We present LeAD, a dual-rate autonomous driving architecture integrating imitation learning-based end-to-end (E2E) frameworks with large language model (LLM) augmentation. The high-frequency E2E subsystem maintains real-time perception-planning-control cycles, while the low-frequency LLM module enhances scenario comprehension through multi-modal perception fusion with HD maps and derives optimal decisions via chain-of-thought (CoT) reasoning when baseline planners encounter capability limitations. Our experimental evaluation in the CARLA Simulator demonstrates LeAD's superior handling of unconventional scenarios, achieving 71 points on Leaderboard V1 benchmark, with a route completion of 93\%.
\end{abstract}

\section{INTRODUCTION}

Autonomous driving systems have witnessed significant advancements in recent years, particularly since the inception of E2E architectures, where deep learning-based models have achieved remarkable performance improvements. However, large-scale open-road deployment of such systems remains infeasible. Beyond challenges like perception limitations and insufficient training data coverage for extreme long-tail scenarios, a critical barrier lies in models' deficient processing capabilities within high-density complex traffic environments and irregular traffic situations\cite{garfinkle2023traffic}. The predominant challenge stems from models' inability to understand scenarios and make human-aligned decisions when interacting with other participants, particularly in interpreting scene semantics and generating logically consistent strategies.

Although functional perception-planning-control pipelines are already mature, existing approaches\cite{bansal2018chauffeurnet,gu2021densetnt} still exhibit critical limitations in complex scenarios: 1) simplified solutions relying on free-space detection for direct trajectory generation, or 2) opaque data-driven models trained on extensive human driving datasets.  These methodologies constrain agents' capacity for semantic comprehension of traffic contexts and interactive reasoning with heterogeneous traffic participants, particularly in negotiating right-of-way conflicts and cooperative navigation scenarios.

To enhance inter-individual interaction capabilities, prior research has predominantly employed game-theoretic\cite{zhou2024game, fang2024cooperative} modeling and optimization-based approaches\cite{nguyen2016driver}, formulating vehicle interactions through mathematical formalisms and handcrafted rule systems. These methods establish predefined interaction protocols for various road scenarios by solving constrained optimization problems that encode traffic regulations and behavioral norms.   However, such rule-driven frameworks face inherent limitations in computational scalability and edge case generalization, primarily manifested in two critical aspects:  1) Exponential computational complexity growth due to the NP-hard nature of multi-agent trajectory optimization, and 2) Fundamental mathematical intractability in modeling edge scenarios characterized by dynamic uncertainties, particularly when confronting non-cooperative human behaviors and unregulated traffic configurations.

The emergence of LLM has introduced novel paradigms for scenario comprehension and interactive reasoning, fundamentally redefining approaches to semantic interpretation and multi-individual coordination within dynamic traffic environments, which has demonstrated remarkable capabilities in scene comprehension and cognitive reasoning\cite{huang2023voxposer}, particularly in three critical dimensions for autonomous driving: 1) contextual interpretation of traffic scenarios\cite{chen2024driving}, 2) knowledge retrieval in complex scenarios\cite{li2024driving} and 3) human-aligned decision logic emulation through symbolic reasoning chains\cite{wei2022chain,sima2023drivelm}. However, a fundamental paradox arises when deploying LLMs: Sustaining their knowledge density and reasoning fidelity inherently restricts operational frequencies to an extremely low level. This latency bottleneck induces critical decision delays, which may lead to serious collision accidents.

\begin{figure}[htbp]
  \centering
  \includegraphics[width=\columnwidth, trim=0cm 11cm 10cm 0cm, clip]{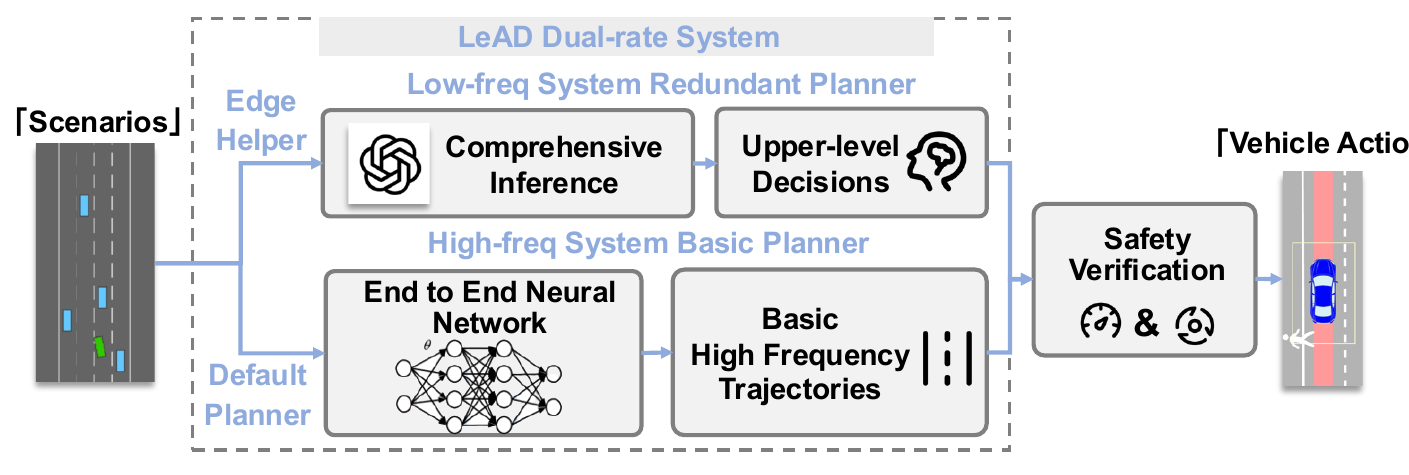}
  \caption{LeAD's Basic Architecture. The proposed architecture integrates LLMs with the E2E framework to enhance autonomous driving systems' capability in handling edge scenarios.}
  \label{introduction}
\end{figure}

We present LeAD, a novel autonomous driving system addressing these scene comprehension challenges by harnessing the exceptional contextual reasoning capabilities of LLMs. The architecture integrates dual decision modules: a high-frequency E2E system for real-time trajectory planning and a redundant LLM-based system for cognitive augmentation. The E2E fundamental system is built upon transformer artitechture\cite{chitta2022transfuser}, working for multi-modality perception and planning to ensure the agent's real-time reaction and security. The LLM subsystem incorporates four core components—natural language encoding of perceptual data, traffic scenario description, chain-of-thought\cite{wei2022chain} reasoning, and safety-constrained decision generation. The encoder transforms multi-modal sensor data into linguistically structured representations, while the scenario parser organizes traffic elements into hierarchical semantic graphs. Through CoT-guided analysis, the LLM deduces optimal actions subsequently converted into executable commands via rule-based safety validators. Dual systems are coordinated by a novel asynchronous coupling mechanism, which enables complementary operation between the high-frequency reactive layer and low-frequency cognitive layer.

This design achieves enhanced edge scenario adaptability without compromising real-time responsiveness. Experimental results on the CARLA Leaderboard demonstrate substantial performance advantages of our method over existing E2E baselines.  Ablation studies further validate the critical role of the LLM subsystem, with improvements in two core metrics: route completion and driving score.

The contributions of this paper are summarized as follows:
\begin{itemize}
    \item LeAD, an innovative autonomous driving system, is proposed, which leverages a LLM for cognitive scene interpretation and human-like logical reasoning, enabling scene-comprehension grounded planning.
    \item A bidirectional natural language encoder-decoder is designed to enable transformation between perception/decision data and linguistic representations, which enhances the LLM's capacity for reliable information comprehension and logical decision-making.
    \item A dual-rate system architecture is presented, which synergistically integrates a real-time-capable E2E framework with a LLM-augmented module possessing scene comprehension and reasoning capabilities, assisting LeAD for its efficient operation, as well as a successful completion of the CARLA Leaderboard autonomous driving closed-loop test.
\end{itemize}

\section{Related Works}

\subsection{Autonomous Driving Planner.}
Recent years have witnessed extensive research on path planning for autonomous vehicles, mainly focusing on three methodological paradigms: data-driven learning approaches, game-theoretic modeling\cite{zhou2024game, fang2024cooperative, liu2024cooperative, liu2024enhancing}, and optimization-based techniques. Among these, data-driven methods and optimization frameworks currently dominate practical implementations. Data-driven approaches, such as reinforcement learning(RL)\cite{chekroun2023gri, chen2021learning, zhang2021end, toromanoff2020end, xu2025tell} (online/offline) and imitation learning(IL)\cite{hu2023planning, Chitta2023PAMI, shao2023reasonnet}, enable policy acquisition from environmental interactions and enormous training datasets respectively. IL demonstrates remarkable adaptability in complex driving conditions through behavioral cloning, while RL achieves powerful policies through billions of interactions with its environment. On the other hand, optimization-based methods\cite{nguyen2016driver, gao2018optimal, li2021optimization} formulate vehicle interactions and trajectory generation as constrained mathematical problems, leveraging efficient solvers to derive provably optimal solutions. Their rigorous mathematical foundations ensure reliable performance under well-defined operational design domains. However, critical limitations persist: learning-based approaches still face critical limitations of interpretability, security and multi-participants scenarios comprehension; optimization frameworks face computational intractability (NP-hard complexity) and long-tail scenario modeling failures.

\subsection{Large Language Models in Autonomous Driving.}
The emergence of LLMs has spurred numerous works\cite{chen2024driving, tian2024drivevlm} in autonomous driving systems. LLMs demonstrate versatile applications spanning human-vehicle interaction, decision interpretability enhancement, and scene understanding. GPT-Driver\cite{mao2023gpt} tried linguistic modeling of driving behaviors, utilizing LLMs to generate navigational trajectories. DriveGPT4\cite{xu2024drivegpt4} incorporates multimodal LLMs into driving tasks, enabling action reasoning through video sequences and textual explanations while addressing user queries. LanguageMPC\cite{sha2025languagempclargelanguagemodels} specializes in complex scenario interpretation and high-level decision inference. LMDrive\cite{shao2024lmdrive} implements language-based end-to-end autonomous driving within closed-loop environments. Nevertheless, current LLM-integrated autonomous driving systems always neglect critical constraints including temporal responsiveness and safety assurance, which are paramount for reliable closed-loop deployment.

In this work, LeAD proposes a novel integration of large language models with end-to-end  autonomous driving systems, synergistically combining real-time operational capabilities with advanced cognitive reasoning to achieve superior decision-making in complex scenarios.

\section{Methodology}
\label{sec:methodology}

\subsection{Overview}
The architectural framework of LeAD is illustrated in Fig. 2. The dual-system decision architecture comprises two parallel subsystems: a high-frequency fast system based on an E2E network and a redundant low-frequency slow system utilizing a LLM, integrated through LeAD's asynchronous coupling mechanism.

\begin{figure*}[t]
  \centering
  \includegraphics[width=\textwidth, trim=1cm 2.5cm 0.5cm 1cm, clip]{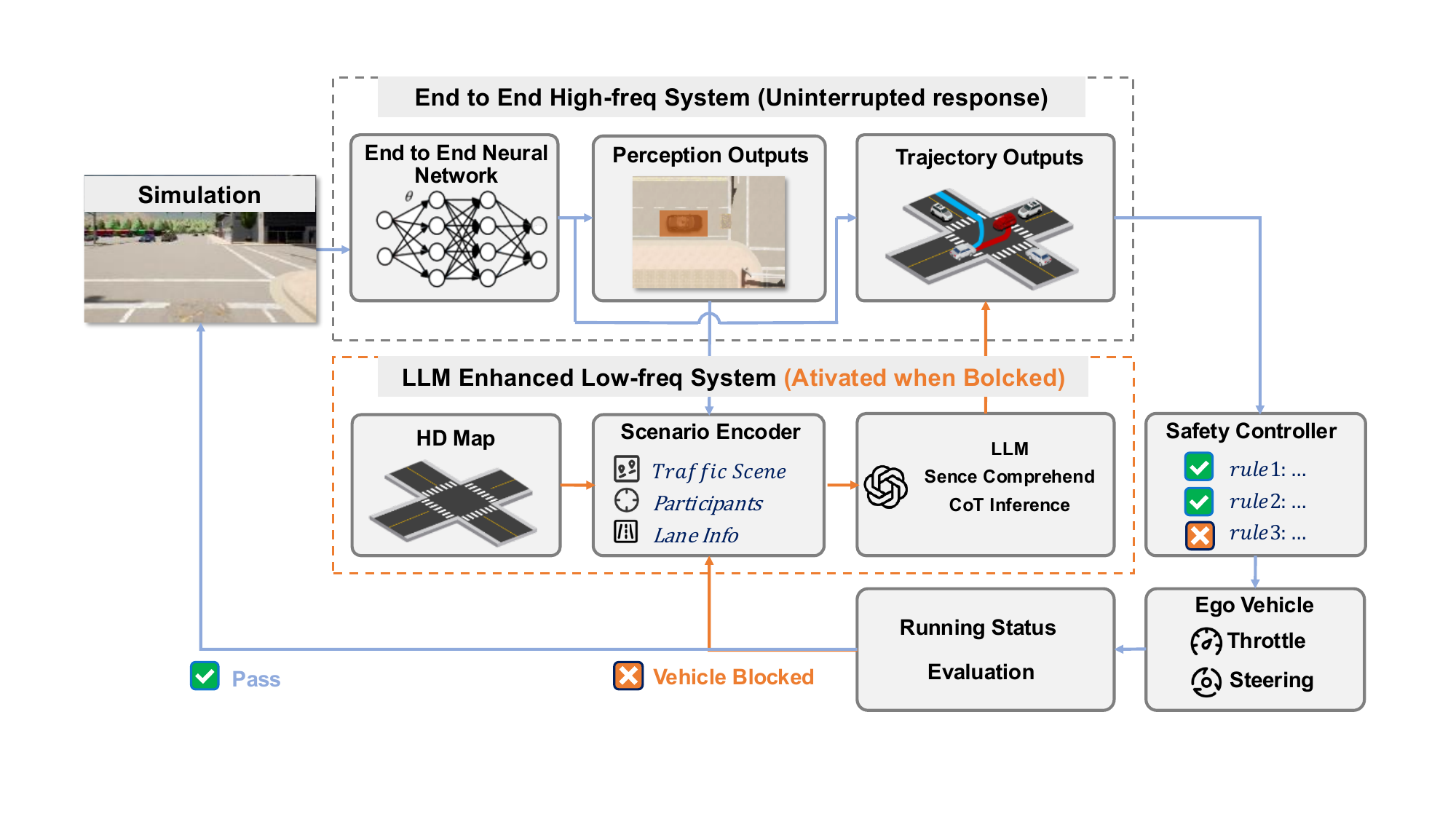}
  \caption{Our dual-rate decision architecture comprises a high-frequency primary system performing real-time object detection and path planning via deep neural networks, and a low-frequency LLM-augmented system enhancing edge case handling through scene understanding and high-level reasoning, with coordinated operation through asynchronous coupling mechanisms.}
  \label{dual_system}
\end{figure*}

Designed for real-time operation, the Fast E2E Subsystem processes multimodal sensor data, then outputs multi-object perception results and generates trajectory waypoints\cite{shao2023safety}, achieving immediate vehicle actuation commands. On the other hand, the LLM Slow Subsystem operates at a low frequency. This slow subsystem performs scene semantics extraction from environmental states, and uses chain-of-thought\cite{wei2022chain} to make strategic decisions for edge scenario resolution planning, leveraging pretrained knowledge to comprehend and handle complex traffic interactions. Finally, LeAD is formed through our asynchronous operation coupling mechanism.



\subsection{Dual-rate System Interaction Mechanism}

LeAD integrates a LLM-based redundant decision-making system with a baseline E2E framework, establishing a dual-rate autonomous driving architecture.     This hybrid system synergizes the LLM's scene understanding and reasoning capabilities with the E2E system's high-frequency responsiveness, enhancing edge case handling capacity while fulfilling real-time operational requirements.

The system primarily operates through the E2E pipeline, which continuously generates trajectory waypoints fed to the base safety controller for immediate vehicle actuation. When encountering edge scenarios: 1)The vehicle initially attempts resolution via the E2E decision stream. 2)The safety controller initiates a protective stop if the scenario remains unresolved. 3)Upon exceeding a predefined waiting threshold, the LLM redundant system activates. 4)The LLM processes current environmental states and perception data to formulate high-level decision, then the vehicle action will be made via the slow system's safety controller.

Furthermore, directly executing decisions decoded from LLM inferences may induce safety risks due to the model's inherent complexity and inability to achieve real-time responsiveness. Therefore, LeAD integrates a safety controller into the slow system decision pipeline. By downgrading safety redundancies from the E2E fast system's safety controller while preserving fundamental collision avoidance constraints, this design ensures sufficient execution freedom for high-level LLM decisions at low speeds.

\subsection{End-to-end Fast Planning System}

\subsubsection{Input and Output Representations.} The system inputs primarily consist of three camera groups, one LiDAR point cloud, and navigation waypoints. The triple-camera setup captures RGB images at front, left-front, and right-front orientations, processed through the RxRx3 image encoding framework. LiDAR point clouds are transformed into histogram representations on a 2D BEV grid\cite{prakash2021multi,ye2023fusionad,liu2023bevfusion,lang2019pointpillars}. Navigation data is encoded in a relative coordinate system for trajectory prediction.

Outputs primarily comprise three components: 1) traffic participants' perception data (relative position, velocity, heading angle, and 2D bounding box dimensions), 2) ego-vehicle trajectory predictions, and 3) traffic sign/intersection detection results derived from the triple-camera streams, enabling compliant traffic signal responses.

\subsubsection{Network Architecture.}
The end-to-end system's deep neural network adopts a ResNet\cite{he2016deep}+Transformer\cite{vaswani2017attention} architecture. Capitalizing on ResNet's exceptional visual task performance and hierarchical feature extraction capabilities, the network achieves multi-level feature abstraction from low-level to high-level representations. Simultaneously, the Transformer enhances inter-feature relationship modeling through self-attention mechanisms, enabling richer semantic expression across hierarchical levels. By synergistically combining these two architectures, the network effectively fuses multi-sensor data (camera images and LiDAR point clouds), significantly improving system perception and prediction capabilities.

Specifically, the ResNet backbone processes input images and pillar-based histograms\cite{prakash2021multi,ye2023fusionad,liu2023bevfusion} through convolutional operations to generate abstract feature maps. These features are fed into the Transformer encoder-decoder structure, where the encoder's multi-head self-attention mechanisms and the decoder's task-specific embeddings/queries jointly generate heterogeneous outputs. Final predictions are produced through dedicated prediction heads.

\begin{figure}[htbp]
  \centering
  \includegraphics[width=\columnwidth, trim=0cm 5cm 0cm 1cm, clip]{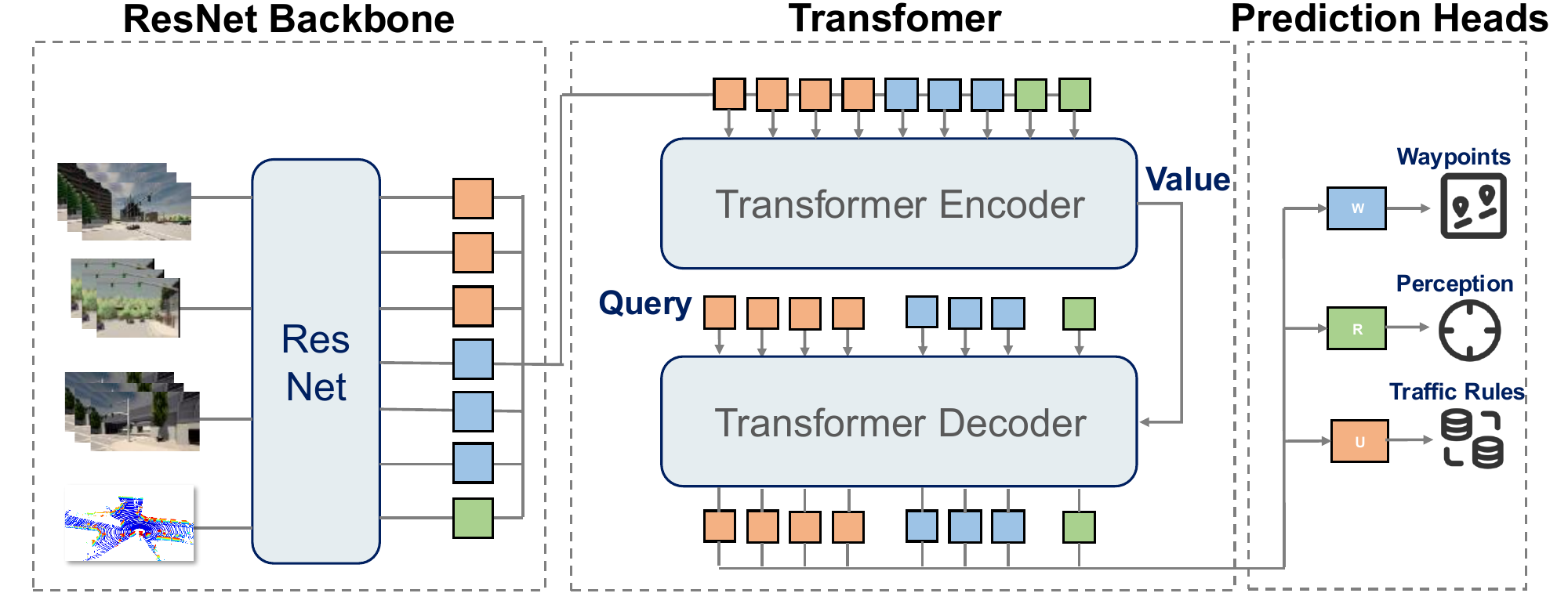}
  \caption{End-to-end system neural network architecture. Building upon previous methodologies, our network architecture integrates ResNet with Transformer modules, processing multimodal inputs and finally generating two principal outputs: multiple perception results and trajectory predictions.}
  \label{network}
\end{figure}

\subsubsection{Vehicle Safety Rules.}
Given the limitations of sparse feedback signals (desired speed and steering angle) for effective neural network training, following the previous study\cite{shao2023safety}, our network design exclusively outputs waypoint predictions and object perception results. To derive executable control commands while enhancing safety, we implement a constraint-based control strategy that integrates safety distance analysis and car-following models.

By projecting objects and waypoints onto a 2D BEV plane (Considering position prediction). The system calculates minimum safe distances through spatial occupancy analysis and applies the intelligent driver model (IDM) to determine desired speed based on real-time safety margins. Steering angle generation employs a preview-based approach, selecting look-ahead waypoints at distance X and transforming their relative coordinates to steering by PID\cite{wang2019path} controller.

Traffic-rule-constraints activate the emergency brake when traffic sign recognition modules detect stop signs or traffic lights with confidence scores exceeding threshold, ensuring compliance with traffic regulations.

\begin{figure*}[t]
  \centering
  \includegraphics[width=\textwidth, trim=0cm 6cm 1.25cm 0cm, clip]{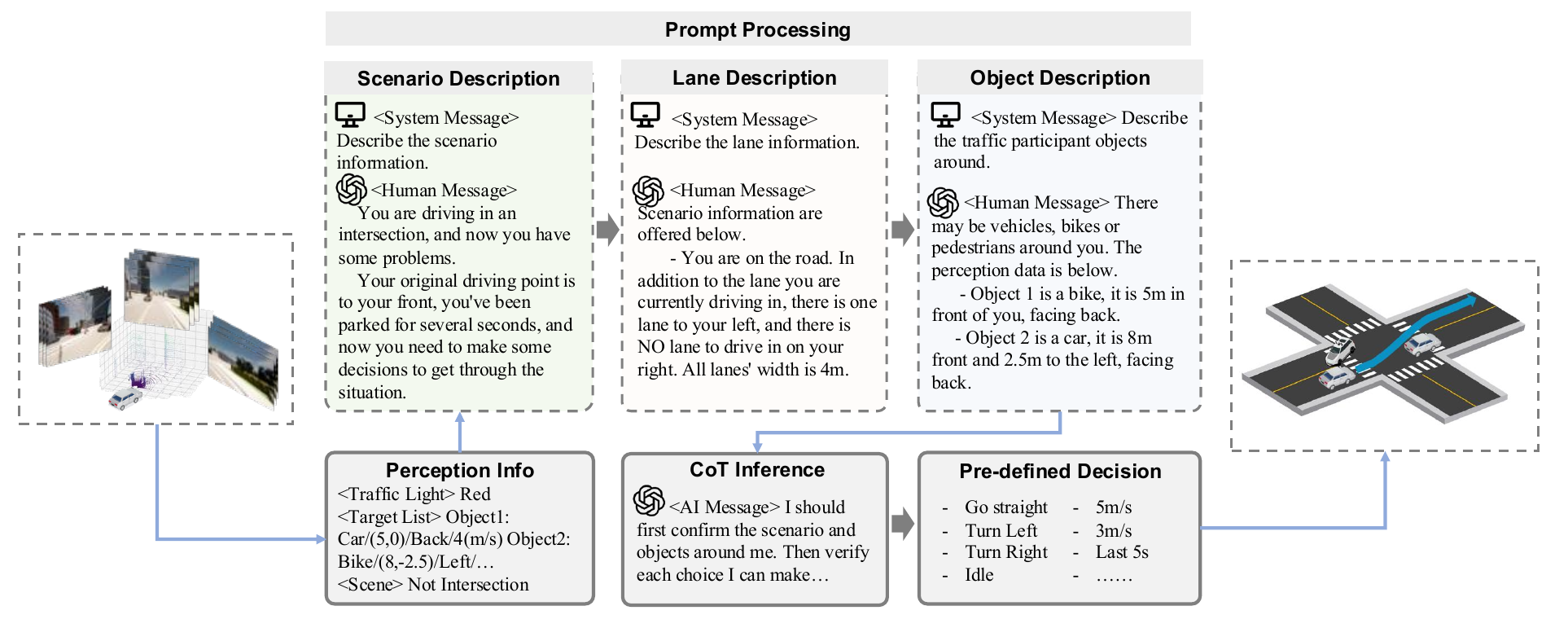}
  \caption{LLM Prompt Pipline in LeAD. The framework implements hierarchical structuring of scene components through tiered classification, guiding the LLM to employ CoT reasoning processes.  This design ensures thorough scene comprehension and logically valid decisions.}
  \label{prompt_fig}
\end{figure*}

\subsection{Perception Description}
To achieve more reliable and precise decision-making with LLMs, in addition to employing higher-performance LLMs, meticulous prompt engineering constitutes another critical determinant of decision quality. Simply inputting raw system outputs as unstructured lists into LLMs inevitably yields suboptimal responses.

Since LLMs cannot directly interface with simulation environments to execute vehicle control, our encoder design processes two distinct input streams. Perception outputs from the E2E system, including detected objects, traffic signals, and partial road features. Lane-level localization data extracted from HD maps, providing ego-vehicle positioning and decision space constraints.
\\
\subsubsection{Road Topology}
The vehicle's operational environment constitutes the primary factor influencing driving decisions.   Given identical perceptual inputs, an ego-vehicle positioned at intersections versus within travel lanes may exhibit fundamentally divergent behavioral strategies. Therefore, the scene description prompts in our system prioritize linguistic characterization of the immediate driving environment through four operational dimensions:
a.Road Context: Determines whether the vehicle resides within a regular lane or intersection area, impacting detour distance estimation.
b.Lane Configuration: Identifies adjacent drivable lanes, influencing lane-change and waiting strategies.
c.Traffic Signals: Detects the presence of stop-indicating traffic lights ahead.
d.Stop Signs: Guides movement strategies in signal-free environments.


\subsubsection{Traffic Participants}
Beyond the vehicle's immediate environment, the perception information of traffic participants constitutes another critical factor influencing driving behavior. However, unlike the end-to-end approach, we exclusively concentrate on key attribute information of each target. This is due to the inherent numerical insensitivity of large language models - directly inputting quantitative measurements such as target dimensions and coordinates would lead to model misinterpretation and consequently erroneous decisions.

Therefore, in designing our prompt engineering, we intentionally attenuate numerical precision in raw perception data and transform it into relatively discrete attribute information, including: 1) target's positional relationship relative to the ego-vehicle , 2) target's heading discretized into 8 cardinal directions, and 3) target's type classification, used for qualitatively judging the target size to determine passability.

\subsection{Inference decision}
To fully leverage the scene comprehension and logical reasoning capabilities of LLM, LeAD structures the CoT-based decision process into three sequential phases.
\\
\subsubsection{Perception Target Analysis}
The target analysis accesses the road occupancy from both temporal and spatial perspectives. The process initiates with describing the driving environment, followed by analyzing each target's potential lane/area occupation. Subsequently, it predicts the targets' probable positions within the forthcoming time window using their heading angles and approximate velocities.

\subsubsection{Decision Feasibility Evaluation}
For all predefined feasible decisions (e.g., "turn right", "HOLD", or "target speed 5 m/s"), this phase systematically analyzes each option's executability and evaluates its operational consequences. The assessment criteria ensure decision optimality through: a) Safety impact prediction; b) Executability in physical space; c) Minimum waiting time.

\subsubsection{Final Decision Synthesis and Decode}
This component converts LLM-generated natural language decisions into executable commands. Through structured prompt design, we enforce formatted output generation with predefined discrete decision options. Each permitted decision (e.g., "lane change left", "decelerate") is mapped to preconfigured trajectory parameters (steering curvature, target speed, execution duration) that are subsequently processed by the safety controller.
\\

Furthermore, to enhance LLMs' contextual understanding and logical decision-making capabilities, the encoder integrates exemplar demonstrations following standardized encoding templates, including a representative driving scenario description and an inference example demonstrating LeAD's CoT\cite{wei2022chain} reasoning process. These CoT templates explicitly articulate reasoning steps.


\section{Simulation Settings}
The simulation tests were conducted using CARLA Simulator v0.9.10\cite{Dosovitskiy17} and the CARLA Leaderboard V1\cite{carla2020leaderboard} framework. The evaluation comprises systematic benchmarking across 26 distinct routes in 7 CARLA-provided towns to evaluate operational capabilities, covering diverse traffic scenarios and urban typologies, including rural lanes, urban arterials, multi-lane boulevards, complex signalized intersections, and unsignalized junctions.

The evaluation protocol strictly prohibits manual interventions during test runs. Persistent immobilization beyond an extended threshold triggers automatic test termination and route failure recording.

The autonomous vehicle will be required to complete predefined test routes from specified origins to destinations. During navigation, the ego-vehicle must maintain collision-free operation while strictly adhering to traffic regulations. Mirroring real-world navigation systems, sparse waypoints are provided as directional references for road selection.

To ensure experimental consistency and comparable performance, we implemented a controlled testing protocol:

\begin{itemize}
    \item Identical origin-destination pairs are maintained across different model evaluations.
    \item Weather/illumination conditions remain fixed.
    \item Traffic scenarios are reproduced through deterministic event triggering when vehicles reach predefined geofenced areas.
\end{itemize}

Our end-to-end neural network was trained and validated on a computational platform equipped with an Intel Core-14700K CPU, NVIDIA GeForce RTX 4080 Super GPU, and 32GB DDR5 RAM. For language model integration, we employed the GPT-4o-mini\cite{openai2024gpt4omini} API in zero-shot mode, with all prompts adhering to the standardized formatting requirements referring to Eureka\cite{ma2023eureka}.

\section{Simulation and Performance Evaluation}

\begin{table*}[t]
\centering
\caption{Performance Rank of CARLA Leaderboard-V1 Test.}
\begin{tabularx}{\textwidth}{@{}c>{\centering\arraybackslash}X>{\centering\arraybackslash}X>{\centering\arraybackslash}X>{\centering\arraybackslash}X>{\centering\arraybackslash}X>{\centering\arraybackslash}X@{}}
\toprule
Rank & Model         & Driving Score & Route Completion & Infraction Penalty & Route Timeouts & Agent Blocked \\ \midrule
1    & LeAD          & 71.96         & 93.43            & 0.76               & 0.06           & 0.04          \\
2    & TF++          & 61.57         & 77.66            & 0.81               & 0.00           & 0.71          \\
3    & Transfuser    & 61.18         & 86.69            & 0.71               & 0.01           & 0.43          \\
4    & TCP(r)        & 58.56         & 83.14            & 0.70               & 19.74          & 50.84         \\
5    & Transfuser(r) & 55.04         & 89.65            & 0.63               & 0.01           & 0.50          \\ \bottomrule
\end{tabularx}

\label{tab:leaderboard}
\end{table*}

\begin{table*}[t]
\centering
\caption{Ablation Study of LLM Module in LeAD.}
\begin{tabularx}{\textwidth}{@{}c>{\centering\arraybackslash}X>{\centering\arraybackslash}X>{\centering\arraybackslash}X>{\centering\arraybackslash}X>{\centering\arraybackslash}X>{\centering\arraybackslash}X@{}}
\toprule
Test & Module\&Map         & Driving Score & Route Completion & Infraction Penalty & Route Timeouts & Agent Blocked \\ \midrule
1    & LeAD\&Town01  & 25.44         & 64.56            & 0.41               & 0.03           & 0.08          \\
2    & AD\&Town01    & 19.67         & 51.31            & 0.42               & 0.14           & 0.01          \\
3    & LeAD\&Town02  & 55.48         & 79.03            & 0.65               & 0.37           & 0.54          \\
4    & AD\&Town02    & 44.49         & 51.27            & 0.88               & 3.38           & 0.00          \\
5    & LeAD\&Town04  & 74.52         & 88.70            & 0.84               & 4.24           & 0.00          \\
6    & AD\&Town04    & 71.27         & 85.47            & 0.84               & 5.31           & 0.00          \\ \bottomrule
\end{tabularx}

\label{tab:ablation}
\end{table*}

\subsection{Baselines}
As shown in Table \ref{tab:leaderboard}, several models are selected from the CARLA Leaderboard V1\cite{carla2020leaderboard} as baselines: TCP\cite{wu2022trajectory} is an end-to-end method with multi-time-step trajectory prediction, based on camera sensor input. Transfuser\cite{chitta2022transfuser} is another end-to-end method based on imitation learning, which uses the attention mechanism to fuse camera and lidar inputs to extract scene information. Transfuser(r) is an improved version of Transfuser. TF++\cite{jaeger2023hidden} a variant of Transfuser, which improves the performance by enhancing the output of trajectories planning.

\subsection{Metrics}
Our evaluation methodology adopts the CARLA Leaderboard V1 metric system\cite{carla_leaderboard_v1_2020}. Performance assessment is conducted through three principal metrics. The evaluation executes individual benchmarking runs for each navigation route across all selected maps, with subsequent score aggregation through route-length-weighted averaging to produce the final performance score.
\begin{itemize}
    \item Driving Score: As the primary evaluation metric, this score represents the product of Route Completion and Penalty Multiplier, scaled to a maximum of 100 points.
    \item Route Completion: Measured as the percentage of successfully navigated route segments (r\%), normalized to a 100-point scale.  Vehicle deviations from drivable areas result in proportional deductions from the completion ratio.
    \item Penalty Multiplier: Initialized at 1.0, this multiplier decreases with each traffic violation until reaching the minimum value of 0.  Violation categories include: a) Collisions with pedestrians, vehicles, or static objects (traffic lights/signs). b) Traffic light violations (entering intersections during red phases). c) Stop sign disregard (failure to perform complete stops)
\end{itemize}
In addition, if the vehicle is driven off the road, the off-road part is included in the uncompleted route.   In other words, the score of route completion is deducted.

\subsection{Performance Evaluation}

Table \ref{tab:leaderboard} presents comparative results between LeAD and four models from the CARLA Leaderboard V1\cite{carla2020leaderboard}. Experimental data indicates that LeAD achieves the 1-st position out of the selected models based on Driving Score, attaining a peak score of 71.96 (Route Completion Rate: 93.43\%, ranking 1-th; Penalty Score: 0.76, ranking 2-th).

\subsection{Ablation Study}

To evaluate the contribution of the integrated LLM-based redundant decision module, we conducted an ablation study by removing the LLM module from the LeAD architecture, retaining only the core end-to-end driving system as the baseline model (AD). Both the full LeAD system and the ablated baseline were evaluated on the Town01-long, Town02-long, and Town04-long benchmarks to enable a performance comparison.

As detailed in Table \ref{tab:ablation}, the baseline system exhibited two predominant failure modes: traffic congestion caused by imprecise trajectory planning and occupancy prediction deviations stemming from velocity estimation errors, both frequently leading to vehicle immobilization or navigation failures. Furthermore, abnormal queuing behaviors of other traffic participants under specific scenarios resulted in traffic flow jams.

The LLM-enhanced system demonstrated substantial operational improvements through contextual reasoning capabilities. By interpreting complex scene semantics and generating high-level behavioral strategies, the LLM auxiliary module reduced average traversal duration by 23.7\% and improved route completion rates by 18.4\% compared to the standalone E2E baseline. These enhancements are attributed to the LLM's capacity for anticipatory conflict resolution and adaptive decision-making under partial observability conditions.

Furthermore, the selected Town01-long, Town02-long, and Town04-long represent three routes with increasing levels of difficulty.  Results show that the performance gains brought by the LLM module become more pronounced as the route complexity increases.  This trend highlights LeAD’s enhanced capacity for decision-making and adaptive response under challenging and dynamic traffic scenarios.

Notably, LeAD performs slightly worse than AD in the Blocked metric. This is primarily because the LLM occasionally chooses to maintain a ``Hold'' decision, causing the vehicle to wait for extended periods and be marked as blocked. However, this conservative behavior helps LeAD avoid collisions or deadlocks, ultimately contributing to better performance in the other metrics.

\subsection{Cases Analysis}
We conducted empirical validation in a representative two-lane bidirectional roadway scenario to demonstrate LeAD's enhanced irregular scene handling capabilities.

\begin{figure}[htbp]
    \centering
    \subfloat[Frame 1]{\includegraphics[width=0.08\textwidth]{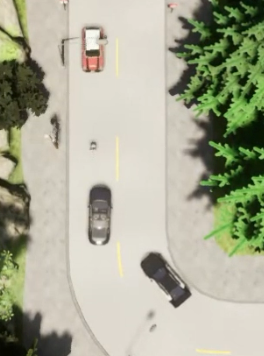}} 
    \hfill
    \subfloat[Frame 2]{\includegraphics[width=0.08\textwidth]{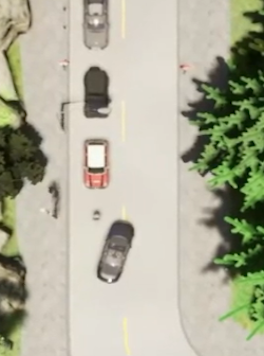}} 
    \hfill
    \subfloat[Frame 3]{\includegraphics[width=0.08\textwidth]{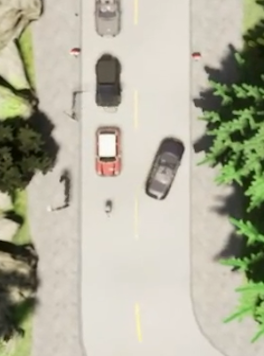}} 
    \hfill
    \subfloat[Frame 4]{\includegraphics[width=0.08\textwidth]{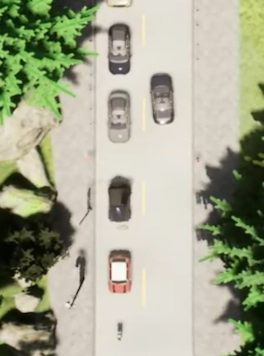}} 
    \hfill
    \subfloat[BEV]{\includegraphics[width=0.08\textwidth]{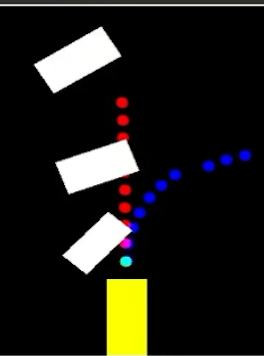}} 
    \\
    \subfloat[LLM Choice Analysis]{\includegraphics[width=0.5\textwidth]{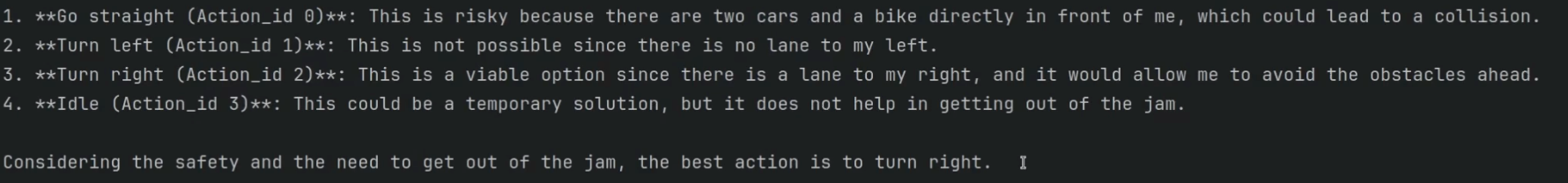}} 
    \caption{A Sample Case. (a)-(d) present sequential video frames depicting vehicle maneuver execution. (e) illustrates trajectory waypoints comparisons, with red markers denoting baseline planner outputs and blue markers indicating LLM-augmented decision trajectories. (f) provides a process visualization of the LLM decision-making chain.}
    \label{fig:llmexample}
\end{figure}

As shown in Fig. \ref{fig:llmexample}, the experiment simulated emergency conditions involving sudden oncoming vehicles and a disabled vehicle in the ego lane. When the baseline E2E planner failed to resolve path obstruction caused by unexpected obstacles, resulting in complete immobilization, the redundant decision system activated through multi-stage safety triggers.  Natural language reasoning generates a ``right lane change'' command after chain-of-thought analysis of traffic regulations and spatial constraints.

\begin{figure}[htbp]
  \centering
  \includegraphics[width=0.4\textwidth, trim=9cm 4.5cm 9cm 0cm, clip]{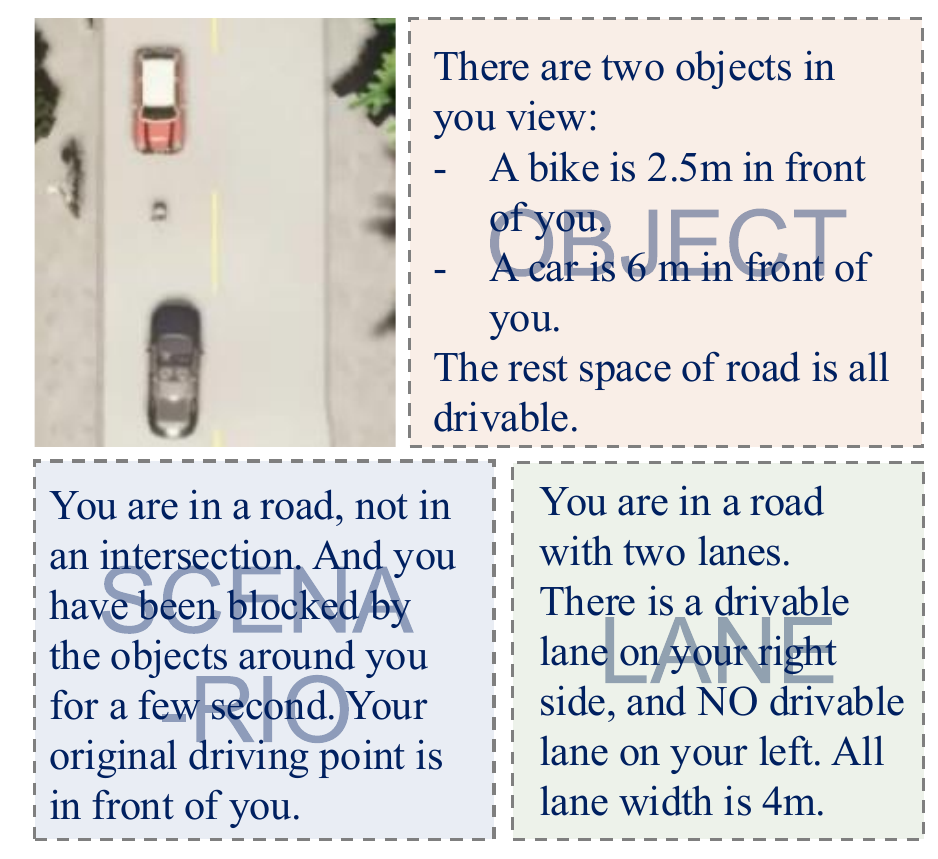}
  \caption{Description of the example case}
  \label{exp_llm}
\end{figure}

\section{CONCLUSIONS}

Enhancing scene semantic comprehension and reasoning capabilities remains a key challenge in autonomous driving. In this work, we propose a dual-rate-system autonomous driving architecture LeAD, that effectively combines real-time E2E planning with LLM-based high-level reasoning. Key bricks also include the use of structured natural language encoding of perception data and a CoT-based decision inference pipeline. Experimental results on complex scenarios demonstrate LeAD's significant performance, which outperforms many strong baselines, achieving a top driving score of 71.96, with a route completion rate of 93.43\%, confirming its solid planning ability and outstanding scene passability.  The proposed framework offers a practical reference for future research on cognitively enhanced decision-making in autonomous systems.

Ablation studies further validate the effectiveness of the LLM-enhanced module: compared to the E2E-only baseline, LeAD improves the average route completion by up to 18.4\% on selected routes, and reduces the average traversal time by 23.7\%, showing the significant contributions of the LLM module to the driving performance in generalization and operational robustness across diverse scenarios.

Future research will focus on expanding LeAD's decision-making sophistication through three directions: 1) refining LLMs' scene interpretation granularity and action specification precision, 2) developing direct waypoint generation interfaces bridging high-level decisions with motion planning, and 3) constructing a specialized dataset aligning natural language reasoning processes with perception-prediction patterns. This dataset can be used for LLM fine-tuning to enable a more human-like, safety-compliant, and efficiency-optimized decision across complex traffic conditions.

\bibliographystyle{IEEEtran}  
\bibliography{reference}

\end{document}